\documentclass[10pt,twocolumn,letterpaper]{article}

\usepackage[pagenumbers]{cvpr} % To force page numbers, e.g. for an arXiv version
\usepackage{multirow}
\usepackage{marvosym}
\newcommand{\TODO}[1]{\textbf{\color{red}[TODO: #1]}}
\renewcommand{\TODO}[1]{}

\definecolor{cvprblue}{rgb}{0.21,0.49,0.74}
\usepackage[pagebackref,breaklinks,colorlinks,allcolors=cvprblue]{hyperref}

\def\paperID{7815} % *** Enter the Paper ID here
\def\confName{CVPR}
\def\confYear{2026}

\title{Disentangled Textual Priors for Diffusion-based Image Super-Resolution}

\author{Lei Jiang \and Xin Liu \and Xinze Tong \and Zhiliang Li \and Jie Liu\textsuperscript{\Letter} \and Jie Tang \and Gangshan Wu\\
State Key Laboratory for Novel Software Technology, Nanjing University, Nanjing 210023, China\\
{\tt\small \{leijiang2024,xinliu2023,xinze\_tong,zhiliang\_li\}@smail.nju.edu.cn} \\ \tt\small \{liujie,tangjie,gswu\}@nju.edu.cn \\
\url{https://github.com/JL6666JL/DTPSR}
}

\begin{document}
\maketitle
\begingroup
\renewcommand\thefootnote{}%
\footnotetext{\Letter:Corresponding author (liujie@nju.edu.cn).}%
\addtocounter{footnote}{-1}%
\endgroup
\begin{abstract}
Image Super-Resolution (SR) aims to reconstruct high-resolution images from degraded low-resolution inputs. While diffusion-based SR methods offer powerful generative capabilities, their performance heavily depends on how semantic priors are structured and integrated into the generation process. Existing approaches often rely on entangled or coarse-grained priors that mix global layout with local details, or conflate structural and textural cues, thereby limiting semantic controllability and interpretability. In this work, we propose DTPSR, a novel diffusion-based SR framework that introduces disentangled textual priors along two complementary dimensions: spatial hierarchy (global \vs local) and frequency semantics (low- \vs high-frequency). By explicitly separating these priors, DTPSR enables the model to simultaneously capture scene-level structure and object-specific details with frequency-aware semantic guidance. The corresponding embeddings are injected via specialized cross-attention modules, forming a progressive generation pipeline that reflects the semantic granularity of visual content—from global layout to fine-grained textures. To support this paradigm, we construct DisText-SR, a large-scale dataset containing approximately 95{,}000 image-text pairs with carefully disentangled global, low-frequency, and high-frequency descriptions. To further enhance controllability and consistency, we adopt a multi-branch classifier-free guidance strategy with frequency-aware negative prompts to suppress hallucinations and semantic drift. Extensive experiments on synthetic and real-world benchmarks show that DTPSR achieves high perceptual quality, competitive fidelity, and strong generalization across diverse degradation scenarios.
\end{abstract}  

\section{Introduction}

Image Super-Resolution (SR), which aims to reconstruct High-Resolution (HR) images from Low-Resolution (LR) inputs, plays a vital role in applications including medical imaging, remote sensing, surveillance, and historical image restoration. Traditional SR methods~\cite{chen2023activating,chen2021pre,zhang2022efficient,dong2014learning,liang2021swinir,zhang2018image} primarily focus on minimizing pixel-wise errors through distortion-based metrics like PSNR and SSIM~\cite{wang2004image}. Although numerically effective, these approaches often fail to produce visually realistic results with fine details.

To overcome these limitations, perceptual-driven SR approaches have emerged, aiming to enhance both visual quality and semantic fidelity. Generative Adversarial Networks (GANs)~\cite{goodfellow2020generative} have been widely adopted to reduce the perceptual gap and produce more plausible reconstructions~\cite{wang2018esrgan}. However, they often suffer from instability and mode collapse. In contrast, diffusion-based generative models—especially latent-space variants~\cite{wang2024exploiting,lin2024diffbir}—have recently demonstrated greater stability and strong capabilities in restoring structure and synthesizing textures, making them promising for perceptual-oriented SR.

Text-to-Image (T2I) diffusion models, such as Stable Diffusion~\cite{rombach2022high}, further extend this potential by leveraging large-scale image–text pretraining. Textual priors serve as native semantic priors, activating the model's generative pathways while offering interpretable and model-aligned guidance—particularly beneficial when the visual input is heavily degraded or lacks sufficient pixel information. Existing text-guided SR methods typically adopt either local tag-based descriptions~\cite{wu2024seesr,tsao2024holisdip} or global sentence-based descriptions~\cite{yang2024pixel,chen2025faithdiff,yu2024scaling}. While both provide useful cues, these strategies tend to focus on either fine-grained detail or global coherence, but rarely capture both. More critically, they rely on frequency-agnostic priors that entangle structure and texture within a single latent representation, limiting controllability and targeted restoration.

To address these issues, we propose a new semantic-guided diffusion framework that leverages disentangled textual priors---explicitly structured along two complementary axes: spatial hierarchy (global \vs local) and frequency semantics (low-frequency \vs high-frequency). Instead of mixing scene layout and object details into a unified embedding, we design separate semantic branches to encode and inject these priors through dedicated cross-attention pathways. Global priors guide the generation of holistic layout, while local priors are further disentangled into low-frequency (\eg, shape, color) and high-frequency (\eg, texture, edges) components that enhance structural integrity and perceptual fidelity, respectively. This design enables semantically aligned and frequency-aware restoration in a progressive manner.

To support this paradigm, we construct DisText-SR, a large-scale dataset consisting of approximately 95{,}000 image-text pairs, where each pair contains a global description of the image and region-wise low- and high-frequency annotations derived from panoptic segmentation. This structured and fine-grained supervision enables precise semantic alignment and facilitates further research in controllable and interpretable SR.

Based on this formulation, we propose DTPSR (Disentangled Textual Priors for Super-Resolution), a diffusion-based framework that systematically integrates structured semantic priors into the generative process. These priors are extracted automatically through a unified pipeline: given an input image, DTPSR employs a segmentation model to obtain object-level regions and a caption generation model to produce corresponding global, low-frequency, and high-frequency descriptions. Through dedicated cross-attention modules, each type of prior is injected via an independent pathway, and a multi-branch classifier-free guidance strategy further improves consistency by suppressing hallucinations through frequency-aware negative prompts.

The main contributions of this work are summarized as follows:

$\bullet$ We propose DTPSR, a diffusion-based super-resolution framework that integrates textual priors disentangled along spatial and frequency dimensions for interpretable and controllable restoration.

$\bullet$ We construct DisText-SR, a new SR dataset with approximately 95{,}000 groups of structured global-local and low-high frequency textual descriptions, enabling fine-grained semantic guidance.

$\bullet$ We introduce a disentangled semantic injection mechanism with separate cross-attention pathways for global, low-frequency, and high-frequency priors, and enhance controllability through multi-branch classifier-free guidance with frequency-aware negative prompts.

$\bullet$ Extensive experiments on synthetic and real-world datasets demonstrate that DTPSR achieves strong perceptual quality, competitive fidelity, and robust generalization.

\section{Related Works}

\subsection{Denoising Diffusion Probabilistic Models}
Diffusion models synthesize data by reversing a noise corruption process. Initially proposed as diffusion probabilistic models~\cite{sohl2015deep}, their practical effectiveness was established by DDPM~\cite{ho2020denoising}, which predicts added noise for stable training and improved sample quality. Subsequent works extended diffusion to the latent space~\cite{rombach2022high}, enabling scalable text-to-image (T2I) generation frameworks such as Imagen~\cite{saharia2022photorealistic}, Stable Diffusion (SD)~\cite{rombach2022high}, and PixelArt-$\alpha$~\cite{chen2023pixart}. These T2I models have proven more capable than GANs in generating realistic, high-quality images~\cite{ramesh2022hierarchical,zhang2023adding,brooks2023instructpix2pix,hertz2022prompt,kawar2023imagic}.

\subsection{Diffusion-Based Super-Resolution}
With the growing success of diffusion models~\cite{mou2024t2i,ramesh2022hierarchical,sahak2023denoising}, several methods have applied them to image super-resolution to better recover textures and high-frequency details~\cite{lin2024diffbir,wang2024exploiting,wang2024sinsr}. StableSR~\cite{wang2024exploiting} introduces learnable SFT layers into pretrained SD to incorporate LR constraints, achieving a trade-off between fidelity and perceptual quality. DiffBIR~\cite{lin2024diffbir} adopts a two-stage strategy: a coarse restoration followed by refinement using Stable Diffusion as a strong prior. However, these methods often neglect textual semantics, missing an opportunity to fully exploit diffusion priors for improved detail and semantic accuracy.

\subsection{Text-Guided Diffusion for Image SR}
To better utilize semantic information, recent works incorporate textual inputs into diffusion-based SR. SeeSR~\cite{wu2024seesr} and DreamClear~\cite{ai2024dreamclear} use short semantic tags or keywords to improve local structure understanding, while PASD~\cite{yang2024pixel}, FaithDiff~\cite{chen2025faithdiff}, and SUPIR~\cite{yu2024scaling} adopt sentence-level captions for global guidance. These methods show that language priors can improve fidelity and realism. In contrast to prior works that rely exclusively on either global or local descriptions, our approach introduces a hierarchical, frequency-aware formulation that disentangles global context and local structure into low- and high-frequency priors—enabling finer control and more faithful restoration.

\section{Method}

\begin{figure}[t]
\centering
\includegraphics[width=0.9\linewidth]{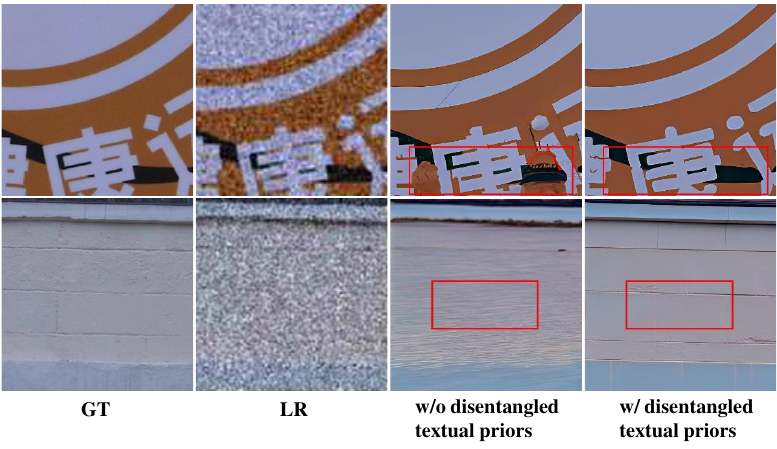}
\caption{Comparison under severe degradation. Without textual priors, the diffusion model suffers from hallucinations, generating human-like artifacts or misinterpreting walls as ocean textures. Incorporating our disentangled textual priors suppresses such errors and enhances semantic consistency.}
\label{fig:severe_degradation}
\end{figure}

\subsection{Motivation and Overview}

Diffusion-based T2I models such as Stable Diffusion~\cite{rombach2022high} have shown strong semantic generation ability, and recent works extend them to SR by incorporating text guidance. However, most rely on coarse cues—such as object tags or scene captions—that do not distinguish structural semantics from fine-grained details, often causing artifacts under severe degradation.

Without disentangled semantic guidance, diffusion models struggle to balance global layout recovery and high-frequency detail synthesis. As shown in \cref{fig:severe_degradation}, coarse or mixed textual cues may lead to hallucinated semantics or misinterpretation of smooth regions as complex textures, revealing a mismatch between unified text conditioning and the multi-scale nature of visual content.

To address this, we propose DTPSR, a diffusion-based SR framework that reformulates the task as semantically guided restoration. We disentangle textual priors along two axes: spatial hierarchy (global \vs local) and frequency semantics (low- \vs high-frequency). Global priors support scene-level structure, while local priors are split into low-frequency cues (\eg, shape, layout) and high-frequency cues (\eg, texture, edges), each injected via dedicated cross-attention pathways for frequency-aware restoration.

To support DTPSR, we construct DisText-SR, a dataset of about 95{,}000 image–text pairs with global captions and region-level low/high-frequency descriptions obtained via panoptic segmentation. This structured supervision enables semantic alignment and controllable generation. During inference, the process is fully automated: given an LR image, segmentation extracts regions, the caption model generates disentangled global and local descriptions, and the diffusion model restores the HR result guided by these priors.

We further introduce a multi-branch classifier-free guidance strategy using frequency-aware negative prompts to suppress hallucinations and enhance controllability. Together, these designs enable DTPSR to deliver strong SR performance under diverse degradations.

\begin{figure*}[t]
\centering
\includegraphics[width=0.9\textwidth]{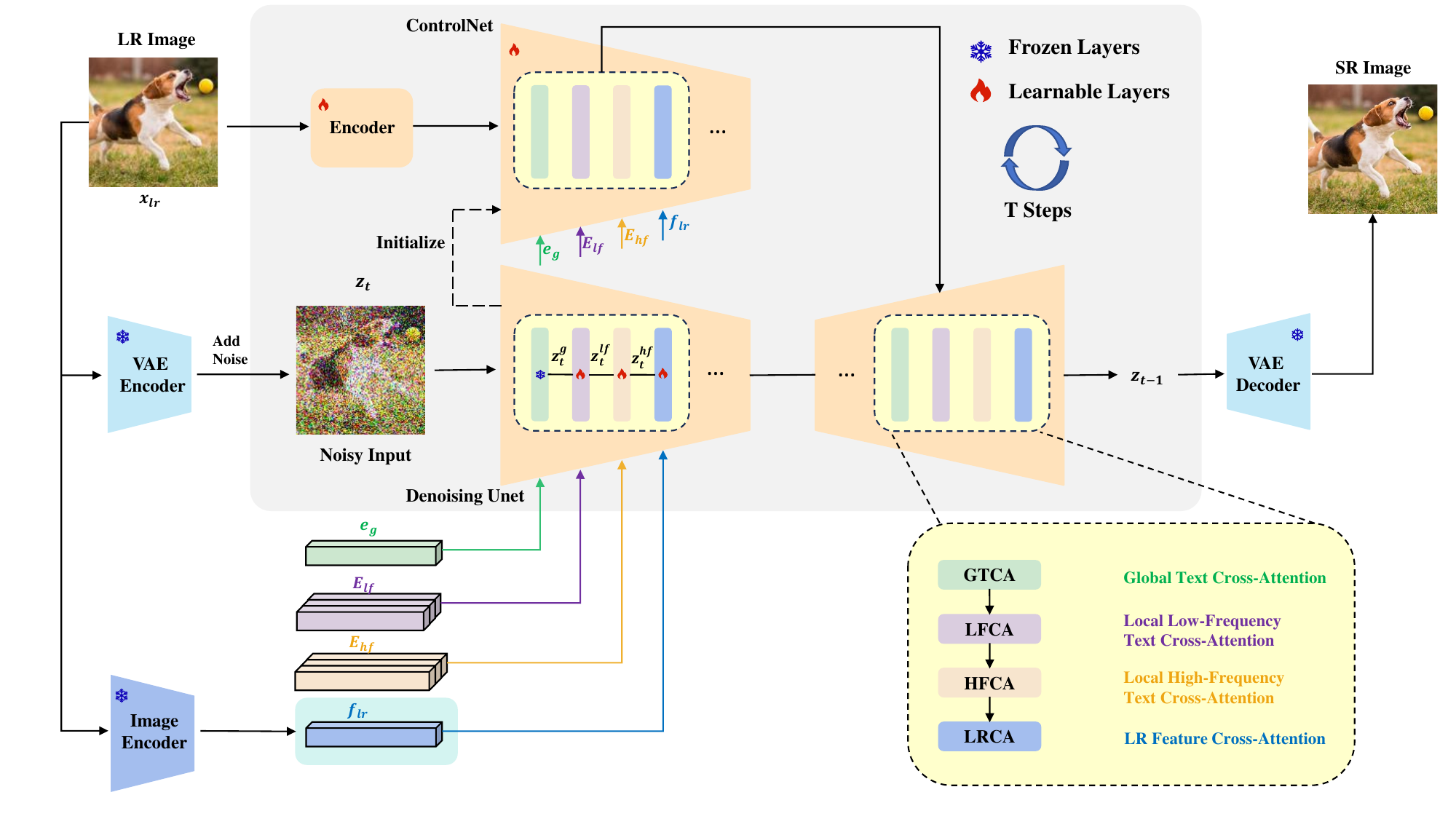} % Reduce the figure size so that it is slightly narrower than the column.
\caption{The overall architecture of DTPSR. Given an LR image, a global prior and local object priors are extracted, where local priors are disentangled into Low-Frequency (LF) descriptions (shape, layout, color) and High-Frequency (HF) descriptions (texture, edges, details). They are encoded into $\mathbf{e}_g$, $\mathbf{E}_{lf}$, and $\mathbf{E}_{hf}$ via CLIP, while the LR image is encoded into $\mathbf{f}_{lr}$ by an image encoder. The diffusion process sequentially updates the latent $\mathbf{z}_t$ through GTCA, LFCA, and HFCA—yielding intermediate representations $\mathbf{z}_t^g$, $\mathbf{z}_t^{lf}$, and $\mathbf{z}_t^{hf}$—and then fuses $\mathbf{f}_{lr}$ via LRCA to produce $\mathbf{z}_{t-1}$, progressively restoring structures and details.}
\label{fig:architecture}
\end{figure*}

\subsection{Design of the DTPSR Framework}

The overall architecture of DTPSR is illustrated in \cref{fig:architecture}. Given a Low-Resolution (LR) image $x_{lr} \in \mathbb{R}^{H \times W \times 3}$, we first map it into the latent space using a pre-trained VAE encoder:
\begin{equation}
z_0 = \text{VAE\_Enc}(x_{lr}).
\label{eq:vae_enc}
\end{equation}

To prepare inputs for training, we apply the forward diffusion for $T$ timesteps, gradually adding Gaussian noise to the latent $z_0$. At timestep $t$, the noisy latent is:
\begin{equation}
z_t = \sqrt{\bar{\alpha}_t} z_0 + \sqrt{1 - \bar{\alpha}_t} \epsilon, \quad \epsilon \sim \mathcal{N}(0, \mathbf{I}),
\label{eq:forward_diffusion}
\end{equation}
where $\bar{\alpha}_t$ denotes the cumulative product of the noise schedule up to $t$.

To reconstruct the clean image during the reverse process, DTPSR employs a structured guidance mechanism that injects different types of disentangled semantic priors—from global layout to fine-grained textures—through specialized cross-attention modules. This enables a coarse-to-fine restoration process that is semantically aligned and frequency-aware, allowing the model to refine structure and perceptual detail progressively and interpretably.

\textbf{Global Text Cross-Attention (GTCA).}  
To establish a strong structural foundation, we introduce a global prior to guide holistic layout generation. A global sentence $c_g$ is encoded via a CLIP text encoder in the first layer, which focuses on restoring holistic scene structure:
\begin{equation}
e_g = \text{CLIP\_TextEnc}(c_g),
\label{eq:clip_global}
\end{equation}
and injected into the noisy latent through a dedicated cross-attention module:

\begin{equation}
z_t^{g} = \text{GTCA}(z_t, e_g).
\label{eq:gtca}
\end{equation}

\textbf{Low-Frequency Cross-Attention (LFCA).}  
After establishing the global layout, the next step enhances structural fidelity at the object level. We encode a set of low-frequency local descriptions $\{c_{lf}^{(i)}\}_{i=1}^{n}$ capturing shape, size, and spatial arrangement:
\begin{equation}
E_{lf} = \left[\text{CLIP\_TextEnc}(c_{lf}^{(1)}), \dots, \text{CLIP\_TextEnc}(c_{lf}^{(n)})\right],
\label{eq:clip_lf}
\end{equation}
and use them to refine $z_t^{g}$ via the proposed LFCA module: 
\begin{equation}
z_t^{lf} = \text{LFCA}(z_t^{g}, E_{lf}).
\label{eq:lfca}
\end{equation}

\textbf{High-Frequency Cross-Attention (HFCA).}  
To capture visual realism and fine textures, we further guide the model with high-frequency semantic cues such as surface details, textures, and edges. They are encoded from a set of fine-grained descriptions $\{c_{hf}^{(j)}\}_{j=1}^{n}$:
\begin{equation}
E_{hf} = \left[\text{CLIP\_TextEnc}(c_{hf}^{(1)}), \dots, \text{CLIP\_TextEnc}(c_{hf}^{(n)})\right],
\label{eq:clip_hf}
\end{equation}
and injected via the proposed HFCA module:
\begin{equation}
z_t^{hf} = \text{HFCA}(z_t^{lf}, E_{hf}).
\label{eq:hfca}
\end{equation}

\begin{figure*}[t]
\centering
\includegraphics[width=0.85\textwidth]{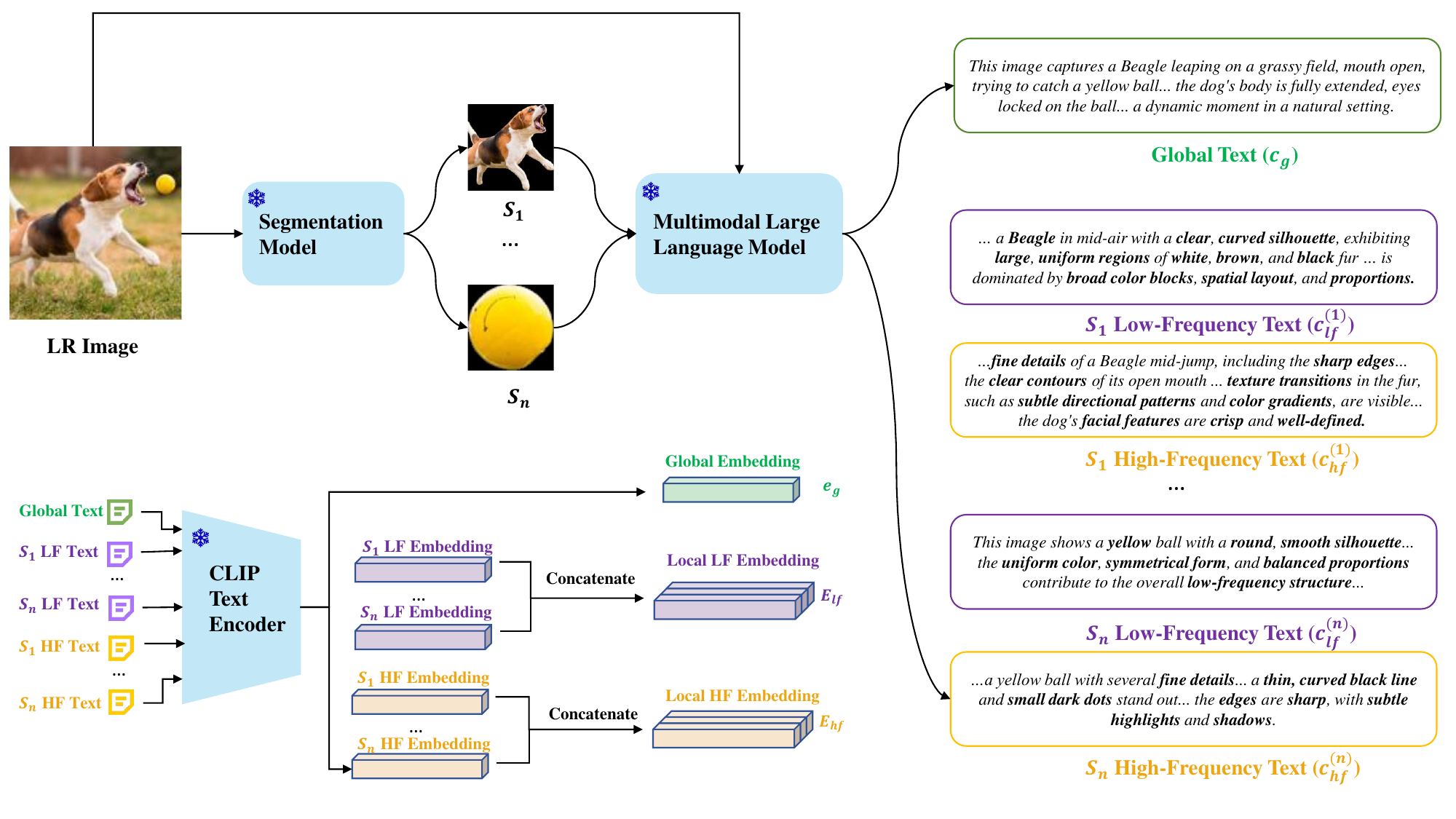} % Reduce the figure size so that it is slightly narrower than the column.
\caption{DisText-SR dataset construction. Given an LR image, we use a segmentation model to extract regions $S_1, S_2, \dots, S_n$, and query a frozen MLLM to generate a global description $c_g$ and per-segment low-frequency ($c_{lf}^{(i)}$) and high-frequency ($c_{hf}^{(i)}$) texts. They are encoded via CLIP into Global, Low-Frequency (LF) and High-Frequency (HF) embeddings for semantic-guided SR.}
\label{fig1}
\end{figure*}

All textual priors—including global, low-frequency, and high-frequency descriptions used in GTCA, LFCA, and HFCA—are illustrated with examples on the right side of \cref{fig1}. The generation procedure for these disentangled textual priors is detailed in the next subsection.

\textbf{Low-Resolution Feature Cross-Attention (LRCA).}  
While textual priors offer strong semantic guidance, they may drift from the original image identity. To anchor the restoration to the input, we introduce an image-specific consistency path. Visual features are extracted using a frozen DAPE encoder~\cite{wu2024seesr}:
\begin{equation}
f_{lr} = \text{DAPE\_Enc}(x_{lr}),
\label{eq:dape}
\end{equation}
and fused into the latent using cross-attention:
\begin{equation}
z_{t-1} = \text{LRCA}(z_t^{hf}, f_{lr}).
\label{eq:lrca}
\end{equation}

Overall, these four modules are designed to operate in a sequential and complementary manner. GTCA provides global scene context; LFCA and HFCA refine the content at increasing levels of detail; and LRCA ensures image-specific consistency. Unlike  methods using flat or entangled cues, our framework aligns priors with the generative trajectory of diffusion, forming a unified pipeline for controllable, interpretable, and semantically faithful restoration.

The complete transformation at timestep $t$ is summarized as:
\begin{equation}
z_t \xrightarrow{\text{GTCA}} z_t^{g} \xrightarrow{\text{LFCA}} z_t^{lf} \xrightarrow{\text{HFCA}} z_t^{hf} \xrightarrow{\text{LRCA}} z_{t-1}.
\label{eq:pipeline}
\end{equation}

To train the model, we optimize a noise prediction network $\epsilon_\theta$ to estimate the injected noise $\epsilon$ based on the noised latent $z_t$ and all guidance inputs:
\begin{equation}
\begin{aligned}
\mathcal{L} = \mathbb{E}_{z_0, z_{lr}, t, c_g, c_{lf}, c_{hf},\ \epsilon \sim \mathcal{N}(0,1)} \Big[\, 
\left\| \epsilon \right. 
\\
\left. - \epsilon_\theta(z_t, z_{lr}, t, c_g, c_{lf}, c_{hf}) \right\|_2^2 
\,\Big].
\end{aligned}
\label{eq:loss}
\end{equation}

\subsection{Construction of the DisText-SR Dataset}

To support fine-grained, frequency-aware super-resolution, we construct a new dataset, DisText-SR, which provides structured and disentangled textual annotations for approximately 95{,}000 images. Each sample contains a global description of the entire image and a set of local region-wise prompts disentangled into low-frequency and high-frequency semantics, as illustrated in \cref{fig1}.

We first apply a pretrained panoptic segmentation model, Mask2Former~\cite{cheng2022masked}, to obtain object-level regions from the input images. These segments, along with the full image, are then passed into a frozen multimodal vision-language model, LLaVA~\cite{liu2023visual}, with carefully crafted prompts designed to elicit hierarchical and frequency-specific descriptions.

For each image, we extract:

$\bullet$ \textit{Global descriptions}, queried with: “Describe the overall scene, including the number of objects, their spatial arrangement, background, and major visual elements.”

$\bullet$ \textit{Low-Frequency descriptions} for each object, using: “Describe the object’s coarse visual structure, including its overall shape, size, dominant colors, and spatial orientation, without mentioning fine details.”

$\bullet$ \textit{High-Frequency descriptions} for each object, using: “Describe the object’s fine-grained visual details, including texture patterns, surface material, edges, and any subtle features or irregularities.”

These texts are encoded using a frozen CLIP~\cite{radford2021learning} text encoder to produce semantic embeddings, which are later injected into our DTPSR framework for aligned and interpretable SR training. To the best of our knowledge, DisText-SR is the first dataset to combine global–local and low–high frequency textual priors at scale, offering a new foundation for studying controllable, semantically grounded image super-resolution.

\subsection{Multi-branch Classifier-Free Guidance}

To enhance semantic controllability, we extend the standard Classifier-Free Guidance (CFG)~\cite{ho2020denoising} into a multi-branch form tailored for disentangled textual priors. Traditional CFG employs a single negative prompt, which is insufficient when handling multiple semantic sources such as global, structural, and textural cues—leading to entangled or inconsistent outputs.

To address this, we introduce a multi-branch CFG strategy with three separate negative prompts: $c_g^{\text{neg}}$ for suppressing incorrect global layout, $c_{lf}^{\text{neg}}$ for low-frequency structural errors, and $c_{hf}^{\text{neg}}$ for high-frequency artifacts. These align with the corresponding positive priors injected via GTCA, LFCA, and HFCA.

This design enables each branch to modulate the generation process independently, enhancing alignment across semantic scales—from global layout to local textures—while reducing hallucinations in a disentangled fashion.

The final noise prediction is computed as:
\begin{equation}
\begin{aligned}
\hat{\epsilon} &= \epsilon_\theta(z_t,z_{lr}, t, c_g, c_{lf}, c_{hf}) \\
\hat{\epsilon}_{\text{neg}} &= \epsilon_\theta(z_t,z_{lr}, t, c_g^{\text{neg}}, c_{lf}^{\text{neg}}, c_{hf}^{\text{neg}}) \\
\tilde{\epsilon} &= \hat{\epsilon} + \lambda_s(\hat{\epsilon} - \hat{\epsilon}_{\text{neg}}),
\end{aligned}
\label{CFG}
\end{equation}
where $\lambda_s$ controls the strength of semantic guidance.

Unlike prior methods~\cite{wu2024seesr,yang2024pixel} using a single negative prompt, our disentangled CFG improves control and semantic fidelity across scales, effectively reducing hallucinations without extra training. More details about the branch-specific negative prompts are provided in the supplementary material (Sec. 2).

\begin{table*}[t]
\centering
\setlength{\tabcolsep}{1mm} 
\caption{Quantitative comparison with state-of-the-art methods on DIV2K-Val, RealSR, and DRealSR datasets. The best and second-best results for each metric are highlighted in \textbf{bold} and \underline{underlined}, respectively.}
\resizebox{\textwidth}{!}{
\begin{tabular}{llccccccccc}
\toprule
\multirow{1}{*}{Dataset} & \multirow{1}{*}{Metrics} & \multicolumn{1}{c}{Real-ESRGAN~\cite{wang2021real}} & \multicolumn{1}{c}{BSRGAN~\cite{zhang2021designing}} & \multicolumn{1}{c}{StableSR~\cite{wang2024exploiting}} & \multicolumn{1}{c}{DiffBIR~\cite{lin2024diffbir}} & \multicolumn{1}{c}{PASD~\cite{yang2024pixel}} & \multicolumn{1}{c}{SeeSR~\cite{wu2024seesr}} & \multicolumn{1}{c}{SUPIR~\cite{yu2024scaling}} & \multicolumn{1}{c}{FaithDiff~\cite{chen2025faithdiff}} & \multicolumn{1}{c}{Ours} \\
\midrule
\multirow{6}{*}{DIV2K-Val~\cite{agustsson2017ntire}} 
& PSNR↑ & 24.12 & \textbf{24.62} & 22.90 & 22.87 & \underline{24.41} & 23.67 & 23.16 & 23.49 & 22.03 \\
& SSIM↑ & \underline{0.6303} & \textbf{0.6311} & 0.5609 & 0.5548 & 0.6252 & 0.6043 & 0.5441 & 0.5817 & 0.5471 \\
& LPIPS↓ & 0.3168 & 0.3390 & \underline{0.3158} & 0.3645 & 0.3794 & 0.3194 & 0.3625 & \textbf{0.3118} & 0.3806 \\
& MUSIQ↑ & 61.29 & 60.52 & 61.06 & \underline{69.29} & 61.27 & 68.67 & 62.59 & 69.18 & \textbf{71.24} \\
& MANIQA↑ & 0.3803 & 0.3502 & 0.3586 & \underline{0.5699} & 0.4042 & 0.5036 & 0.5224 & 0.4309 & \textbf{0.5866} \\
& CLIPIQA↑ & 0.5212 & 0.5311 & 0.6174 & \underline{0.7156} & 0.5568 & 0.6935 & 0.7040 & 0.6463 & \textbf{0.7549} \\
\midrule
\multirow{6}{*}{RealSR~\cite{cai2019toward}} 
& PSNR↑ & 25.56 & \underline{26.38} & 23.84 & 24.65 & \textbf{26.63} & 25.15 & 25.04 & 25.23 & 22.46 \\
& SSIM↑ & 0.7571 & \underline{0.7651} & 0.7050 & 0.6392 & \textbf{0.7660} & 0.7210 & 0.6709 & 0.7069 & 0.6214 \\
& LPIPS↓ & \underline{0.2688} & \textbf{0.2656} & 0.2969 & 0.3667 & 0.2871 & 0.3007 & 0.3715 & 0.2872 & 0.3796 \\
& MUSIQ↑ & 60.15 & 63.28 & 58.12 & 68.99 & 59.99 & \underline{69.82} & 58.51 & 68.86 & \textbf{71.84} \\
& MANIQA↑ & 0.3788 & 0.3758 & 0.3451 & \underline{0.5589} & 0.3939 & 0.5437 & 0.4429 & 0.4644 & \textbf{0.6021} \\
& CLIPIQA↑ & 0.4559 & 0.5114 & 0.5707 & \underline{0.7030} & 0.4855 & 0.6701 & 0.6357 & 0.6126 & \textbf{0.7278} \\
\midrule
\multirow{6}{*}{DRealSR~\cite{wei2020component}} 
& PSNR↑ & 28.61 & \underline{28.66} & 26.33 & 22.87 & \textbf{29.17} & 28.07 & 26.93 & 27.20 & 25.83 \\
& SSIM↑ & \underline{0.8051} & \textbf{0.8151} & 0.7093 & 0.5548 & 0.7954 & 0.7684 & 0.6709 & 0.7102 & 0.7005 \\
& LPIPS↓ & \textbf{0.2819} & \underline{0.2929} & 0.3486 & 0.3645 & 0.3176 & 0.3174 & 0.4310 & 0.3540 & 0.3953 \\
& MUSIQ↑ & 54.28 & 56.80 & 53.42 & \underline{69.18} & 50.42 & 65.09 & 54.16 & 66.30 & \textbf{69.24} \\
& MANIQA↑ & 0.3440 & 0.3354 & 0.3400 & \underline{0.5699} & 0.3685 & 0.5128 & 0.4144 & 0.4526 & \textbf{0.6011} \\
& CLIPIQA↑ & 0.4518 & 0.5126 & 0.6021 & \underline{0.7156} & 0.5059 & 0.6911 & 0.6307 & 0.6335 & \textbf{0.7640} \\
\bottomrule
\end{tabular}
}
\label{tab:quantitative}
\end{table*}

\section{Experiments}
\subsection{Experimental Setting}
\textbf{Datasets.} We follow the training setup of SeeSR~\cite{wu2024seesr}, using the LSDIR~\cite{li2023lsdir} dataset and the first 10,000 images from FFHQ~\cite{karras2019style}, and generate approximately 95{,}000 LR-HR image pairs via the Real-ESRGAN~\cite{wang2021real} degradation pipeline to construct our DisText-SR dataset. For evaluation, we test on both synthetic and real-world datasets, including DIV2K~\cite{agustsson2017ntire}, RealSR~\cite{cai2019toward}, and DRealSR~\cite{wei2020component}. Following the protocol of StableSR~\cite{wang2024exploiting}, the DIV2K validation set is cropped into 3,000 patches and degraded using Real-ESRGAN. For RealSR and DRealSR, center cropping is used. All LR inputs are resized to 128×128 and HR targets to 512×512.\\
\textbf{Implementation Details.} We adopt the SD-2-base model as our base T2I model. To extract LR embeddings for attention layers, we integrate the pretrained DAPE encoder~\cite{wu2024seesr}. For the segmentation module, we adopt Mask2Former~\cite{cheng2022masked} with a Swin-L~\cite{liu2021swin} backbone, trained on the COCO~\cite{lin2014microsoft} dataset. For training, we utilize an AdamW~\cite{loshchilov2017fixing} optimizer to train our model for 110,000 iterations. The batch size and the learning rate are set to 32 and $5\times10^{-5}$. The training process is conducted on 4 NVIDIA A800 GPUs. For inference, we adopt DDPM sampling with 50 timesteps. The guidance scale $\lambda_s$ in Eq. \eqref{CFG} is set to 7.0. Additional implementation details, including optimizer settings and textual conditioning scales, can be found in the supplementary material.\\
\textbf{Evaluation metrics.} We evaluate our method using standard and widely recognized image quality metrics. For measuring image fidelity, PSNR and SSIM~\cite{wang2004image} are computed in the YCbCr color space to reflect pixel-level similarity. To assess perceptual quality, we incorporate both reference-based and no-reference evaluation protocols, including LPIPS~\cite{zhang2018unreasonable}, MUSIQ~\cite{ke2021musiq}, MANIQA~\cite{yang2022maniqa}, and CLIP-IQA~\cite{wang2023exploring}, providing a comprehensive and diverse set of indicators for evaluating visual fidelity and realism.

\begin{figure*}[t]
\centering
\includegraphics[width=\textwidth]{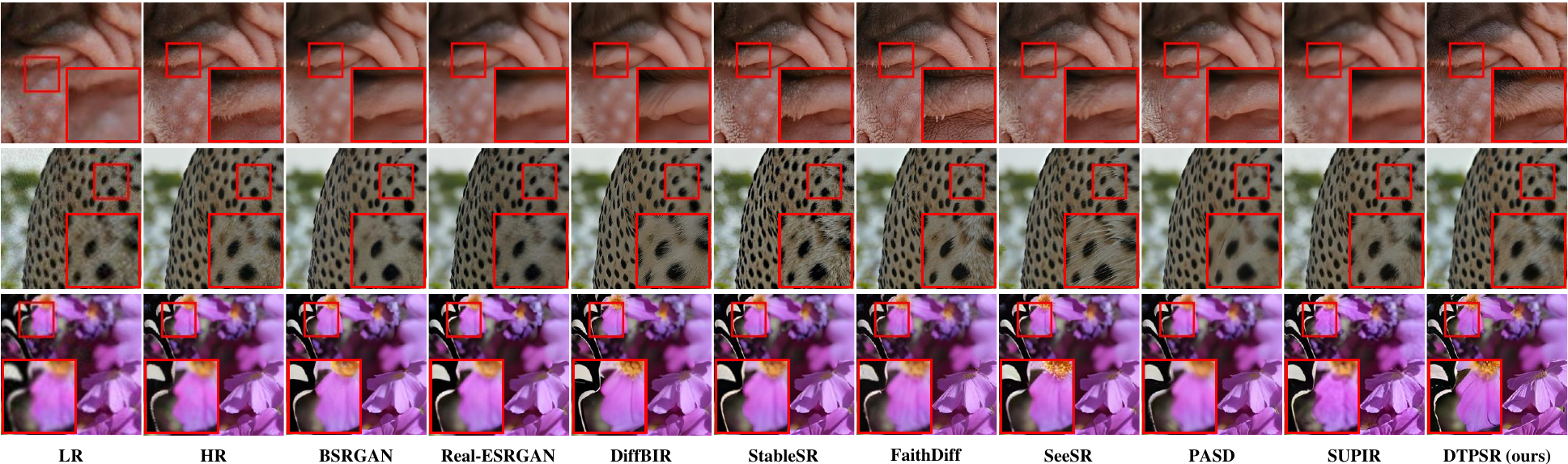}
\caption{Qualitative comparison with representative SR methods. Our DTPSR reconstructs sharper textures and more semantically aligned details, especially under complex degradations, compared to both GAN-based (\eg, BSRGAN, Real-ESRGAN) and diffusion-based (\eg, FaithDiff, SUPIR) approaches. Zoom in for better visual comparison. More qualitative comparisons can be found in Sec. 4 of the supplementary material.}
\label{fig:qualitative_compare}
\end{figure*}

\subsection{Comparison with State-of-the-Arts}
\textbf{Quantitative Comparisons.} As shown in \cref{tab:quantitative}, DTPSR achieves consistently strong perceptual performance across both synthetic and real-world datasets. On the synthetic dataset DIV2K-Val, it attains the highest scores on all no-reference perceptual metrics—MUSIQ (71.24), MANIQA (0.5866), and CLIP-IQA (0.7549)—surpassing the second-best method by clear margins and demonstrating its ability to recover visually plausible details under controlled degradation. On the real-world datasets RealSR and DRealSR, DTPSR again obtains the best MUSIQ (71.84 / 69.24), MANIQA (0.6021 / 0.6011), and CLIP-IQA (0.7278 / 0.7640), highlighting its robustness and generalization to real-world degradations. While GAN-based methods (\eg, BSRGAN, Real-ESRGAN) often yield higher PSNR/SSIM due to distortion-oriented objectives, they lag behind on perceptual metrics. Diffusion-based approaches (\eg, FaithDiff, DiffBIR) improve realism but still trail DTPSR. Overall, the results verify that disentangled textual priors enable DTPSR to achieve superior perceptual super-resolution.

DTPSR does not achieve the highest full-reference scores due to the perception-distortion tradeoff~\cite{blau2018perception}: enhancing perceptual quality often increases pixel-wise distortion. Our method emphasizes realistic texture restoration via disentangled semantic guidance, thus achieving superior performance on no-reference metrics (\eg, MUSIQ, MANIQA, CLIP-IQA). Nevertheless, DTPSR still maintains competitive PSNR, SSIM and LPIPS, demonstrating its ability to balance fidelity and perceptual quality.\\
\textbf{Qualitative Comparisons.} \cref{fig:qualitative_compare} presents qualitative comparisons with several representative methods. GAN-based approaches such as BSRGAN and Real-ESRGAN tend to generate over-smoothed outputs with missing structural details, while diffusion-based models like FaithDiff and SUPIR produce relatively sharper textures but often suffer from spatial inconsistencies, artifacts, or hallucinations. In the first and second rows, our DTPSR reconstructs more realistic and spatially coherent textures, particularly in challenging regions like facial contours and surface materials, where other methods introduce noise or distortion. In the third row, DTPSR delivers noticeably clearer results, preserving edges and fine structures that appear blurred or degraded in other outputs. These visual comparisons highlight the effectiveness of our pixel-to-frequency semantic guidance and disentangled priors in producing perceptually faithful and detail-preserving images.

\subsection{Computational Efficiency}

Although DTPSR includes upstream modules, its overall computation remains moderate. 
\cref{tab:efficiency} compares parameter count and runtime with representative text-guided SR approaches. We restrict this comparison to methods using caption generation models, as non-caption methods differ substantially in design and scale, and caption-guided SR has become a popular approach with generally stronger perceptual performance. For runtime, we report the average per-image inference time on DRealSR ($128\times128 \rightarrow 512\times512$). DTPSR remains efficient due to three factors: a lightweight caption model (LLaVA-7B), segmentation adding only 1.6\% overhead, and processing only the top-3 largest segments. These choices preserve efficiency without sacrificing perceptual quality.

\begin{table}[t]
\centering
\scriptsize
\caption{Comparison of model complexity and runtime efficiency on the DRealSR dataset. Inference time is reported as the average per-image runtime.}
\label{tab:efficiency}
\resizebox{\columnwidth}{!}{
\begin{tabular}{lccc}
\toprule
Method & Params & Steps & Inference Time \\
\midrule
SUPIR~\cite{yu2024scaling} & 17.8B & 100 & 17.25s \\
FaithDiff~\cite{chen2025faithdiff} & 15.6B & 20 & 13.20s \\
DTPSR (Ours) & 10.5B & 50 & 14.94s \\
\bottomrule
\end{tabular}
}
\end{table}

\subsection{Ablation Study}

\begin{table}[t]
\centering
\caption{Ablation study on the effect of global and local textual priors on the DRealSR dataset.}
\resizebox{\columnwidth}{!}{
\begin{tabular}{c|c|c|c|c|c}
\hline
Exp & Global & Local & MANIQA$\uparrow$ & CLIP-IQA$\uparrow$ & MUSIQ$\uparrow$ \\
\hline
3-1 & $\times$ & $\times$ & 0.5271 & 0.7064 & 67.48 \\
3-2 & $\times$ & $\checkmark$ & 0.5851 & 0.7471 & 68.86 \\
3-3 & $\checkmark$ & $\times$ & 0.5394 & 0.7211 & 67.80 \\
3-4 & $\checkmark$ & $\checkmark$ & \textbf{0.6011} & \textbf{0.7640} & \textbf{69.24} \\
\hline
\end{tabular}
}
\label{tab:ablation-global-local}
\end{table}

\begin{table}[t]
\centering
\caption{Ablation study on the effect of frequency-aware textual priors on the DRealSR dataset.}
\resizebox{\columnwidth}{!}{
\begin{tabular}{c|c|c|c|c|c}
\hline
Exp & HF & LF & MANIQA$\uparrow$ & CLIP-IQA$\uparrow$ & MUSIQ$\uparrow$ \\
\hline
4-1 & $\times$ & $\checkmark$ & 0.5850 & 0.7544 & 68.96 \\
4-2 & $\checkmark$ & $\times$ & 0.5495 & 0.7301 & 68.08 \\
4-3 & $\checkmark$ & $\checkmark$ & \textbf{0.6011} & \textbf{0.7640} & \textbf{69.24} \\
\hline
\end{tabular}
}
\label{tab:ablation-frequency}
\end{table}

\textbf{Effect of Global-Local Textual Priors.} As shown in \cref{tab:ablation-global-local}, using only local priors (Exp 3-2) significantly improves perceptual metrics compared to the baseline without any priors (Exp 3-1), demonstrating their effectiveness in refining object-level details. Global priors alone (Exp 3-3) offer moderate gains by enhancing structural consistency. Combining both (Exp 3-4) achieves the best overall performance, confirming that our global-local textual guidance effectively balances scene layout and fine-grained restoration in perceptual-oriented super-resolution.\\
\textbf{Effect of Frequency-Aware Textual Priors.} We first analyze the individual and joint contributions of low- and high-frequency textual priors to perceptual-oriented super-resolution performance. As shown in \cref{tab:ablation-frequency}, using only low-frequency priors (Exp 4-1) achieves better perceptual quality than using only high-frequency ones (Exp 4-2), as structural guidance helps restore global layout and object coherence. However, relying solely on low-frequency cues limits the model's ability to recover fine textures. The combination of both priors (Exp 4-3) achieves the best results across all metrics, demonstrating that the dual-branch design—separately attending to structure and detail—yields more comprehensive and balanced restoration.\\
\textbf{Disentangled \vs Mixed Frequency-Aware Learning.} 
Beyond evaluating the presence of frequency-aware priors, we further explore how these priors are utilized within the model architecture. 
As shown in \cref{tab:freq_disentangle}, our disentangled strategy (Exp 5-2) injects low- and high-frequency cues into separate LFCA and HFCA branches, enabling specialized processing of structure and texture. 
In contrast, the mixed strategy (Exp 5-1) uses a unified embedding where low- and high-frequency semantics appear in a single sentence and are injected into both LFCA and HFCA. This simpler design weakens semantic separation and reduces interpretability. The higher perceptual scores of the disentangled variant highlight the benefit of separating frequency-aware cues for better control and realism.\\
\textbf{Effect of Multi-Branch Negative Prompts.} We conduct an ablation to assess the effectiveness of multi-branch negative prompting in classifier-free guidance. As shown in \cref{tab:ablation-negative-prompts}, removing negative prompts (Exp~6-1) results in weaker perceptual quality due to the lack of semantic suppression. Adding a single generic prompt (Exp~6-2) improves CLIP-IQA and MUSIQ but lacks the granularity for precise control. Our full method (Exp~6-3), with separate negative prompts for global, low-frequency, and high-frequency branches, achieves the best perceptual results. These findings confirm that multi-branch prompting improves both global semantic alignment and local texture fidelity by enabling disentangled, branch-specific suppression of inconsistent or noisy signals.\\
\textbf{Robustness to Upstream Errors.} DTPSR remains robust to imperfections in upstream modules, including segmentation and textual description models. Even with inaccurate segmentation, the model achieves reliable image–text alignment as long as textual priors are correct, since diffusion mainly relies on semantic guidance. Thus, robustness largely depends on handling noisy textual inputs. We simulate corrupted descriptions by randomly replacing part of the words with ``None''. As shown in \cref{tab:robustness}, this corruption causes only a slight performance drop, and DTPSR still surpasses other methods, showing that its disentangled semantic priors remain effective under imperfect inputs.\\
\textbf{Additional Ablation Studies.} We conduct several extended experiments to further analyze the effect of prior injection order and guidance scale, as well as to provide more quantitative and qualitative evidence on RealSR and DRealSR datasets. These additional ablation studies are included in the supplementary material (Secs. 5).

\begin{table}[t]
\centering
\caption{Comparison of frequency-mixed and frequency-disentangled learning on the DRealSR dataset.}
\resizebox{\columnwidth}{!}{
\begin{tabular}{c|c|c|c|c}
\hline
Exp & Type & MANIQA$\uparrow$ & CLIP-IQA$\uparrow$ & MUSIQ$\uparrow$ \\
\hline
5-1 & Mixed & 0.5947 & 0.7527 & 69.05 \\
5-2 & Disentangled & \textbf{0.6011} & \textbf{0.7640} & \textbf{69.24} \\
\hline
\end{tabular}
}
\label{tab:freq_disentangle}
\end{table}

\begin{table}[t]
\centering
\caption{Ablation study on multi-branch classifier-free guidance (CFG) on the DRealSR dataset.}
\resizebox{\columnwidth}{!}{
\begin{tabular}{c|c|c|c|c}
\hline
Exp & CFG Strategy & MANIQA$\uparrow$ & CLIP-IQA$\uparrow$ & MUSIQ$\uparrow$ \\
\hline
6-1 & None & 0.5995 & 0.6992 & 66.73 \\
6-2 & Single & 0.5776 & 0.7499 & 67.87 \\
6-3 & Multi (Ours) & \textbf{0.6011} & \textbf{0.7640} & \textbf{69.24} \\
\hline
\end{tabular}
}
\label{tab:ablation-negative-prompts}
\end{table}

\begin{table}[t]
\centering
\scriptsize
\caption{Robustness comparison under textual corruption on the DRealSR dataset. DTPSR-C denotes the variant with randomly corrupted textual descriptions (``None'' substitution). Despite artificial textual errors, DTPSR remains superior across perceptual metrics.}
\label{tab:robustness}
\resizebox{\columnwidth}{!}{
\begin{tabular}{lccc}
\toprule
Method & MUSIQ↑ & MANIQA↑ & CLIP-IQA↑ \\
\midrule
DiffBIR~\cite{lin2024diffbir} & 68.86 & 0.5589 & 0.7030 \\
PASD~\cite{yang2024pixel} & 59.99 & 0.3939 & 0.4855 \\
SeeSR~\cite{wu2024seesr} & 69.82 & 0.5437 & 0.6701 \\
SUPIR~\cite{yu2024scaling} & 58.51 & 0.4429 & 0.6357 \\
FaithDiff~\cite{chen2025faithdiff} & 68.99 & 0.4644 & 0.6126 \\
\midrule
DTPSR-C & \underline{71.64} & \underline{0.5855} & \underline{0.7061} \\
DTPSR & \textbf{71.84} & \textbf{0.6021} & \textbf{0.7278} \\
\bottomrule
\end{tabular}
}
\end{table}

\section{Conclusion}
We propose DTPSR, a diffusion-based super-resolution framework guided by disentangled textual priors over global–local and low–high frequency dimensions, supported by the large-scale DisText-SR dataset. Using dedicated cross-attention and a multi-branch guidance strategy, DTPSR enables progressive, interpretable restoration with strong perceptual quality across diverse degradations. Remaining challenges include dependence on upstream segmentation and caption models. Future work will investigate adaptive prompt correction, tighter integration with upstream modules, and more efficient diffusion backbones.

\section*{Acknowledgements}
This work was supported by the National Natural Science Foundation of China (Grant No. 62402211), and we appreciate this support for enabling the research.

{
    \small
    \bibliographystyle{ieeenat_fullname}
    \bibliography{main}
}

% WARNING: do not forget to delete the supplementary pages from your submission 

\end{document}